\documentclass[conference, a4paper]{IEEEtran}
\IEEEoverridecommandlockouts
\usepackage{cite}
\usepackage{amsmath,amssymb,amsfonts}
\usepackage{algorithmic}
\usepackage{graphicx}
\usepackage{textcomp}
\usepackage{xcolor}
\usepackage{cite}
\usepackage{booktabs}
\usepackage{multirow}
\usepackage{bm}
\def\BibTeX{{\rm B\kern-.05em{\sc i\kern-.025em b}\kern-.08em
    T\kern-.1667em\lower.7ex\hbox{E}\kern-.125emX}}
\renewcommand\IEEEkeywordsname{Keywords}
\begin{document}

\title{Point Cloud Denoising and Outlier Detection with Local Geometric Structure by Dynamic Graph CNN\\
}

\makeatletter
\newcommand{\linebreakand}{%
  \end{@IEEEauthorhalign}
  \hfill\mbox{}\par
  \mbox{}\hfill\begin{@IEEEauthorhalign}
}
\makeatother

\author{\IEEEauthorblockN{Kosuke Nakayama}
\IEEEauthorblockA{\textit{Graduate School of Fundamental} \\
\textit{Science and Engineering} \\
\textit{Waseda University}\\
Tokyo, Japan \\
kosuke0013@fuji.waseda.jp}
\and
\IEEEauthorblockN{Hiroto Fukuta}
\IEEEauthorblockA{\textit{Graduate School of Fundamental} \\
\textit{Science and Engineering} \\
\textit{Waseda University}\\
Tokyo, Japan \\
taketomohiro@akane.waseda.jp}
\and
\IEEEauthorblockN{Hiroshi Watanabe}
\IEEEauthorblockA{\textit{Graduate School of Fundamental} \\
\textit{Science and Engineering} \\
\textit{Waseda University}\\
Tokyo, Japan \\
hiroshi.watanabe@waseda.jp}
}

\maketitle

\begin{abstract}
The digitalization of society is rapidly developing toward the realization of the digital twin and metaverse. In particular, point clouds are attracting attention as a media format for 3D space. Point cloud data is contaminated with noise and outliers due to measurement errors. Therefore, denoising and outlier detection are necessary for point cloud processing. Among them, \textit{PointCleanNet} is an effective method for point cloud denoising and outlier detection. However, it does not consider the local geometric structure of the patch. We solve this problem by applying two types of graph convolutional layer designed based on the \textit{Dynamic Graph CNN}. Experimental results show that the proposed methods outperform the conventional method in AUPR, which indicates outlier detection accuracy, and Chamfer Distance, which indicates denoising accuracy.
\end{abstract}
\vspace{\baselineskip}

\section{Introduction}
Digital technology is rapidly progressing in various fields toward the realization of the digital twin and metaverse. In particular, point clouds are attracting attention as a format for representing three-dimensional space. Point cloud data can be acquired with laser scanner. However, it contains outliers and noise due to instrument limitations. Noisy data reduces the accuracy of later processing such as object detection, segmentation, and reconstruction. Therefore, denoising and outlier detection are important technologies that form the basis of point cloud processing.

In general, there are statistical methods for point cloud denoising and outlier detection. One is bilateral filtering \cite{b1} based on the distance between each point and a plane fitted to the neighboring points. The other is moving least square \cite{b2} that projects each point onto a surface fitted to the neighborhood points. However, these two methods cannot handle all planes, surfaces, and edges and have the problem of excessive smoothing.

To address this limitation, deep learning methods for point cloud denoising and outlier detection have been developed in the last few years. In particular, \textit{PointCleanNet} \cite{b3} is an effective deep learning-based method. However, it does not consider the local geometric structure of the patch. Therefore, it is difficult to capture noise and outliers.

To solve this problem, we apply two types of graph convolutional layers based on \textit{Dynamic Graph CNN} \cite{b4}. The first is to dynamically construct a neighborhood graph in each layer. The second is to propagate the neighborhood search results in low-dimensional space to the backward layers. These proposed methods solve the problem of not considering the relationship between points in a patch. Experimental results confirm that the proposed methods outperform the conventional method in AUPR, which indicates outlier detection accuracy, and Chamfer Distance, which indicates denoising accuracy.

\section{Related Works}
\subsection{PointCleanNet}
\textit{PointCleanNet} is a point cloud denoising and outlier detection method using deep learning. It consists of two stages: \textit{1) Outlier Detector} for outlier removal and \textit{2) Denoiser} for noise correction. Specifically, the outlier detector classifies outliers and discards them from the original point cloud. Next, the denoiser estimates correction vectors that project the noisy points onto the original clean surface. The method is efficient and robust to various levels of outliers and noise. It is also very easy to incorporate into existing shape processing pipelines because of its simplicity and versatility. However, there are problems with remaining outliers and unnatural distortion of planes and surfaces. This is because the local geometric structure within the patch is not considered in the input point cloud.

\subsection{Dynamic Graph Convolutional Neural Networks}
\textit{Dynamic Graph CNN} is a point cloud classification and segmentation method using deep learning. In the process, the nodes of the graph represent points and edges are established based on pairwise distances between points. This structure effectively transforms point cloud data into a graph representation. In particular, by defining convolution operations on dynamic graphs, local geometric features can be learned. The topology of the graph is used to adaptively determine the receptive field of each point, allowing the user to focus on local neighborhood information that is important for understanding the 3D shape. In addition, mathematical analysis to optimize graph convolutional networks is discussed \cite{b5}.

\begin{figure*}[t]
\centerline{\includegraphics[width=\linewidth]{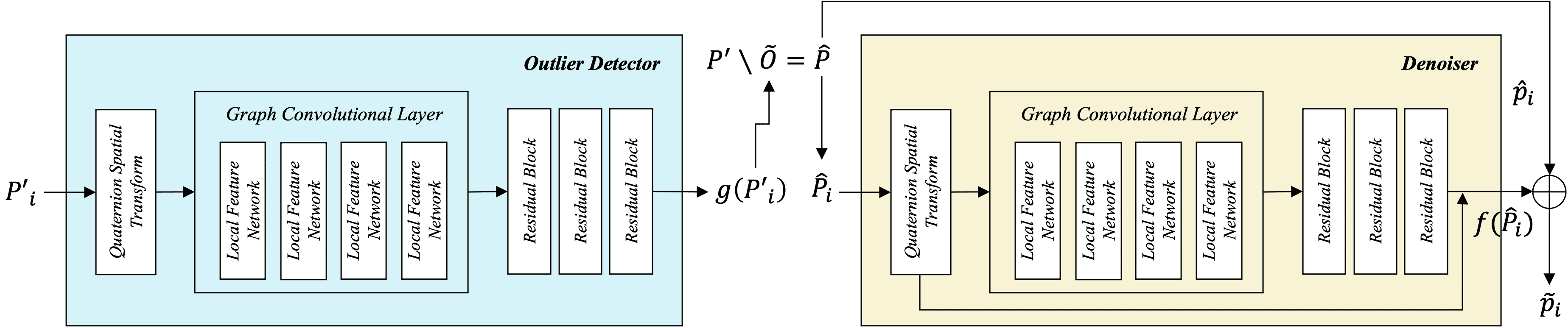}}
\caption{The architecture of the proposed method: Outlier Detector and Denoiser.}
\label{fig}
\end{figure*}

\section{Proposed Method}
\subsection{Architecture}
The architecture consists of two stages: \textit{1) Outlier Detector} for outlier removal and \textit{2) Denoiser} for noise correction. It is shown in Fig. 1. It consists of quaternion spatial transform, graph convolutional layer, and residual block. In particular, the graph convolutional layer consists of four local feature networks. In this network, local neighborhood graphs are constructed by \textit{k}-nearest neighbor search. Then, convolution and addition operations using them are used to generate edge features that can represent relationships between points. They allow the calculation to reflect the local geometric structure information while maintaining the global shape information of the patch.

Quaternion spatial transform is a network that outputs a rotation matrix using quaternions to rotate the patch. It is effective for translational invariance and implicitly learns the rotation transformation to be robust to outliers and noise. Residual block is a network that adds skip connections to the conventional structure. It can be effective in addressing the vanishing gradient problem.

\subsection{Outlier Detector}
The outlier detector takes a local patch ${\textit{P}'}_\textit{i}$ of outlier points ${\textit{P}}$ as input and outputs the outlier estimation probability $\textit{g}\,(\textit{P}'_\textit{i})$ for each point in the patch. After that, an outlier is determined based on the set threshold value. Points determined to be outliers are added to the outlier set $\tilde{\textit{O}}$. The outlier determination is given by
\begin{align}
\textit{If}\:\:{\tilde{\textit{o}}}_{\textit{i}} > \textit{Threshold} \::\: \tilde{\textit{o}}_{\textit{i}}\in \tilde{\textit{O}}.
\end{align}

Then, point cloud $\hat{\textit{P}}$ with outliers removed is given by
\begin{align}
\hat{\textit{P}}=\textit{P}'\:{\backslash}\:\tilde{\textit{O}}.
\end{align}

For the loss, we used the manhattan distance $\textit{L}_{\textit{o}}$ between the estimated outlier label $\tilde{\textit{o}}_{\textit{i}}$ and the correct outlier label $\textit{o}_{\textit{i}}$. The loss function of outlier detector is given by
\begin{align}
\textit{L}_{\textit{o}}(\,\tilde{\textit{p}}_{\textit{i}},\textit{p}_{\textit{i}}) = \| \,\tilde{\textit{o}}_{\textit{i}} - \textit{o}_{\textit{i}} \,\|_{1}.
\end{align}

\subsection{Denoiser}
The denoiser takes as input the local patch $\hat{\textit{P}}_\textit{i}$ of noisy points $\hat{\textit{P}}$ after removing outliers, and outputs the noise correction vector $\textit{f}\,(\hat{\textit{P}}_\textit{i})$ for each point. The denoised smooth point cloud $\tilde{\textit{P}}$ is given by
\begin{align}
\tilde{\textit{P}}=\hat{\textit{P}}\,+\,\textit{f}\:(\hat{\textit{P}}).
\end{align}

For the loss function, we used $\textit{L}_{\alpha}$, a combination of two types of losses $\textit{L}_{s}$ and $\textit{L}_{r}$. $\textit{L}_{s}$ is the square of the L2 distance between each point after denoising and the nearest point in the neighborhood centered at the correct data point corresponding to that point. On the other hand, $\textit{L}_{r}$ is the square of the euclidean distance between the point and the farthest point. These two were weighted by $\alpha$, $\textit{L}_{\alpha}$. These loss functions are represented by
\begin{align}
\textit{L}_{\textit{s}}(\,\tilde{\textit{p}}_{\textit{i}},\textit{P}_{\tilde{\textit{p}}_{\textit{i}}}) &= \underset{\textit{p}_{\textit{j}}\in\textit{P}_{\tilde{\textit{p}}_{\textit{i}}}}{\rm{min}}\|\, \tilde{\textit{p}}_{\textit{i}} - \textit{p}_{\textit{j}} \,\|^2_{2},\\
\textit{L}_{\textit{r}}(\,\tilde{\textit{p}}_{\textit{i}},\textit{P}_{\tilde{\textit{p}}_{\textit{i}}}) &= \underset{\textit{p}_{\textit{j}}\in\textit{P}_{\tilde{\textit{p}}_{\textit{i}}}}{\rm{max}}\|\, \tilde{\textit{p}}_{\textit{i}} - \textit{p}_{\textit{j}} \,\|^2_{2},\\
\textit{L}_{\alpha}(\,\tilde{\textit{p}}_{\textit{i}},\textit{P}_{\tilde{\textit{p}}_{\textit{i}}}) &= {\alpha}{\textit{L}_{\textit{s}}}\,+\,(1-{\alpha})\,{\textit{L}_{\textit{r}}}.
\end{align}

\subsection{Difference of Two Graph Convolutional Layers}
We introduced two types of graph convolution layers. They are shown in Fig. 2. In the first method, based on the normal \textit{Dynamic Graph CNN}. The \textit{k}-neighborhood graph is not fixed, and the graph is dynamically updated after each local feature network. This allows a detailed analysis of the local structure of the neighborhood. In the second method, the graph was designed based on the optimization of graph convolution. The graph is fixed, and the \textit{k}-neighborhood graph in low-dimensional space is also reflected in each local feature network. This is expected to improve the computational speed compared to the usual graph convolution.

\begin{figure}[h]
\centerline{\includegraphics[width=\linewidth]{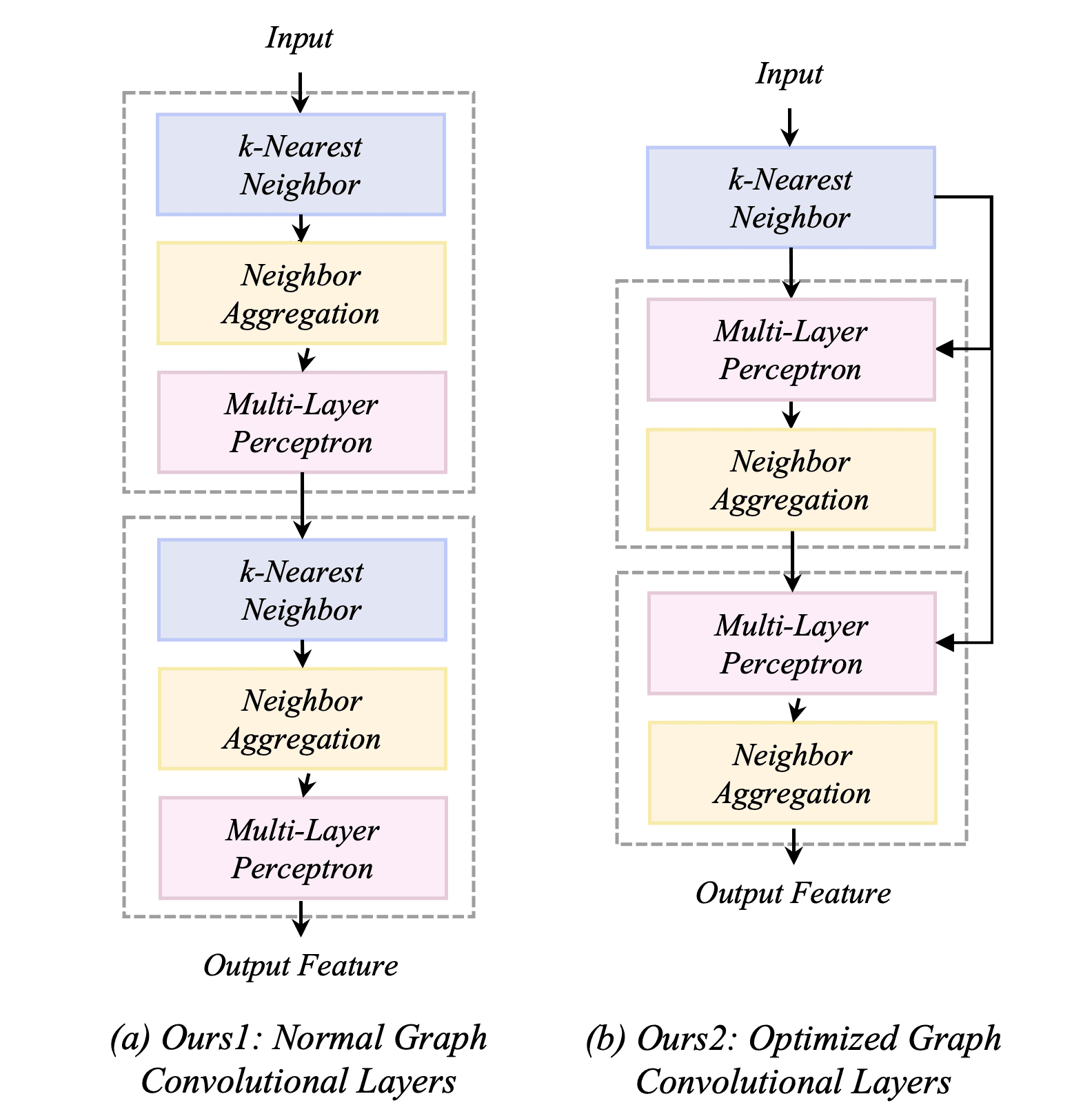}}
\caption{Comparison of two graph convolutional layers. (a) Nomal graph convolution of point clouds. (b) Optimized graph convolution of point clouds.}
\label{fig}
\end{figure}

\section{Experiment}
\subsection{Datasets}
For the outlier detector, the \textit{PointCleanNet} outlier dataset consisting of $28$ different point clouds with $140000$ points for each shape is used. For the denoiser, the \textit{PointCleanNet} noise dataset consisting of $28$ different point clouds with $100000$ points for each shape is used.

\subsection{Experimental Parameters}
For the outlier detector, the outlier threshold is set to $0.5$ and the number of \textit{k}-nearest neighbor searches to $16$. The parameters used in the training were batch size 16, learning rate $10^{-4}$, number of epochs $800$, and initial values of network weights were initialized by He \cite{b6}.
For the denoiser, The $\alpha$ in the loss function $\textit{L}_{\alpha}$ is $0.99$ and the \textit{k}-nearest neighbor search is set to $16$. The parameters used for training were a batch size of $16$, a learning rate of $10^{-8}$, an epoch count of $800$, and a uniform random value of $(-0.001, 0.001)$ for the initial network weights.

\subsection{Evaluation Method}
The average AUPR was used for comparison of outlier detection accuracy. This is because the point cloud of the data set has a bias in the number of outlier and nonoutlier points; the AUPR is a curve representing the change in the reproducibility-fitness ratio of the test results and the study of the area of curvilinearization. The area value ranges from 0 to 1, with larger values indicating higher prediction accuracy.
The average value of Chamfer Distance was used for comparison of denoising accuracy. The similarity of point clouds can be measured by evaluating them against each other from both predicted and ground truth, with lower values indicating higher denoising accuracy. The concept of Chamfer Distance is shown in Fig. 3. Chamfer Distance(CD) is given by
\begin{align}
CD=\frac{1}{\tilde{\textit{P}}}\sum_{\textit{p}_\textit{i} \in \tilde{\textit{P}}}\min_{\textit{p}_\textit{j} \in \textit{P}}\|\textit{p}_\textit{i}-\textit{p}_\textit{j}\|_2^2+\frac{1}{\textit{P}}\sum_{\textit{p}_\textit{j} \in \textit{P}}\min_{\textit{p}_\textit{i} \in \tilde{\textit{P}}}\|\textit{p}_\textit{j}-\textit{p}_\textit{i}\|^2_2.
\end{align}

\begin{figure}[h]
\centerline{\includegraphics[width=\linewidth]{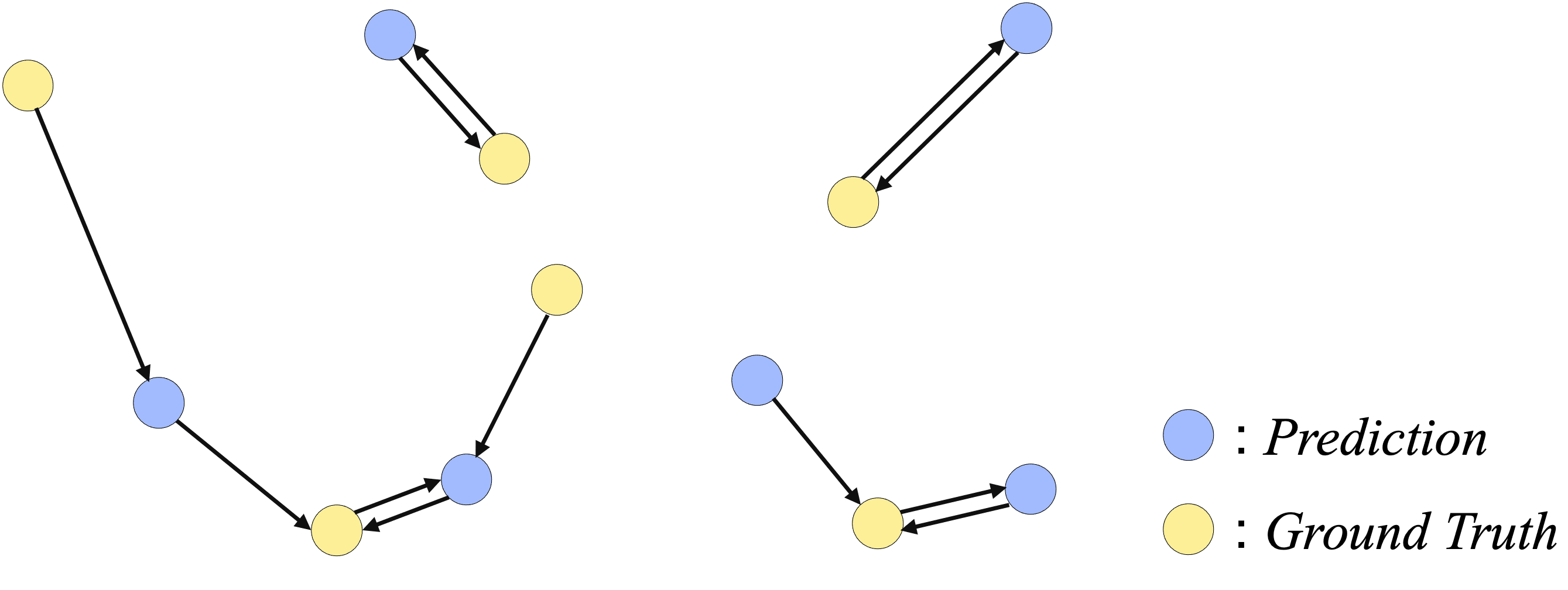}}
\caption{The concept of Chamfer Distance.}
\label{fig}
\end{figure}

\subsection{Accuracy Evaluation}
\textit{1) Outlier Detectoion}: The results of outlier detection accuracy by AUPR are shown in Table I. It shows that "Ours$1$" has the best performance. It outperforms the \textit{PointCleanNet} score by $0.048$ when the noise level is $1.0$.

\textit{2) Denoising}: The results of denoising accuracy by Chamfer Distance are shown in Table I\hspace{-1.2pt}I. It shows that "Ours$1$" has the best performance. It outperforms the \textit{PointCleanNet} score by $0.0035$ when the noise level is $1.5$.

\subsection{Calculation Speed Evaluation}
The comparison of computational speed of the proposed outlier detector and denoiser is also shown in Table I\hspace{-1.2pt}I\hspace{-1.2pt}I. This shows that the performance of the "Ours$2$" is superior.
In particular, the outlier detector can compute each point in $0.36\times10^{-4}$ seconds. Thus, it can be seen that $1,400,000$ points can be computed $50.4$ seconds faster. Also, the denoiser can calculate each point in $0.31\times10^{-4}$ seconds. Thus, it can compute $1,000,000$ points $31.0$ seconds faster.

\begin{table}[h]
\caption{Results of The Outlier Detection Accuracy by AUPR \\ 
with Outlier Detector}
\begin{center}
\renewcommand{\arraystretch}{1.12}
\scalebox{1.1}[1.1] {
\begin{tabular}{c|cccc}
\toprule
\multirow{2}{*}{\textit{Model}} & \multicolumn{4}{c}{\textit{Gaussian Noise Level}}\\
{} & $0$\% & $1.0$\% & $1.5$\% & $2.5$\% \\
 \midrule
 \textit{PointCleanNet} & $0.957$ &  $0.858$ & $0.781$ & $0.659$ \\
 \midrule
 {\textit{Ours$1$}  (\textit{k}\,=$16$)} & \bm{$0.972$} &  \bm{$0.906$} & \bm{$ 0.821$}& \bm{$0.670$} \\
 \midrule
 {\textit{Ours$2$}  (\textit{k}\,=$16$)} & $0.969$ &  $0.902$ & $0.805$ & $0.665$ \\
\bottomrule
\end{tabular}
}
\label{tab1}
\end{center}
\end{table}
\vspace{-1.3mm}
\begin{table}[h]
\caption{Results of The Denoising Accuracy by Chamfer Distance \\ with Denoiser}
\begin{center}
\renewcommand{\arraystretch}{1.12}
\scalebox{1.1}[1.1] {
\begin{tabular}{c|ccc}
\toprule
\multirow{2}{*}{\textit{Model}} & \multicolumn{3}{c}{\textit{Gaussian Noise Level}}\\
{} & $1.0$\% & $1.5$\% & $2.5$\% \\
\midrule
 \textit{PointCleanNet} & $0.0123$ &  $0.0224$ & $0.137$ \\
 \midrule
 {\textit{Ours$1$}  (\textit{k}\,=$16$)} & \bm{$0.0109$} &  \bm{$0.0189$} & \bm{$0.127$} \\
 \midrule
 {\textit{Ours$2$}  (\textit{k}\,=$16$)} & $0.0111$ &  $0.0208$ & $0.134$ \\
\bottomrule
\end{tabular}
}
\label{tab2}
\end{center}
\end{table}
\begin{table}[h]
\caption{Comparison of Computation Time Per Point \\for Outlier Detector and Denoiser}
\begin{center}
\renewcommand{\arraystretch}{1.12}
\scalebox{1.1}[1.1] {
\begin{tabular}{c|cc}
\toprule
{\textit{Model}} & {\textit{Outlier Detector} (\textit{s})} & {\textit{Denoiser} (\textit{s})}\\
 \midrule
 {\textit{Ours$1$}  (\textit{k}\,=$16$)} & $2.57\times10^{-4}$ &  $2.38\times10^{-4}$\\
 \midrule
 {\textit{Ours$2$}  (\textit{k}\,=$16$)} & \bm{$2.21\times10^{-4}$} &  \bm{$2.07\times10^{-4}$} \\
\bottomrule
\end{tabular}
}
\label{tab3}
\end{center}
\end{table}
\vspace{-1.3mm}
\begin{table}[h]
\caption{Results of The Overall Model Performance}
\begin{center}
\renewcommand{\arraystretch}{1.12}
\scalebox{1.1}[1.1] {
\begin{tabular}{c|cc}
\toprule
{\textit{Model}} & {\textit{Chamfer Distance}} & {\textit{Time} (\textit{s})} \\
\midrule
 {\textit{PointCleanNet}} & $1.56\times10^{-4}$ & {$-$} \\
 \midrule
 {\textit{Ours$1$}  (\textit{k}\,=$16$)} & \bm{$1.03\times10^{-4}$}& $5.02\times10^{-4}$ \\
 \midrule
 {\textit{Ours$2$}  (\textit{k}\,=$16$)} & $1.04\times10^{-4}$ & \bm{$4.35\times10^{-4}$} \\
\bottomrule
\end{tabular}
}
\label{tab4}
\end{center}
\end{table}

\begin{figure*}[h]
\centerline{\includegraphics[width=\linewidth]{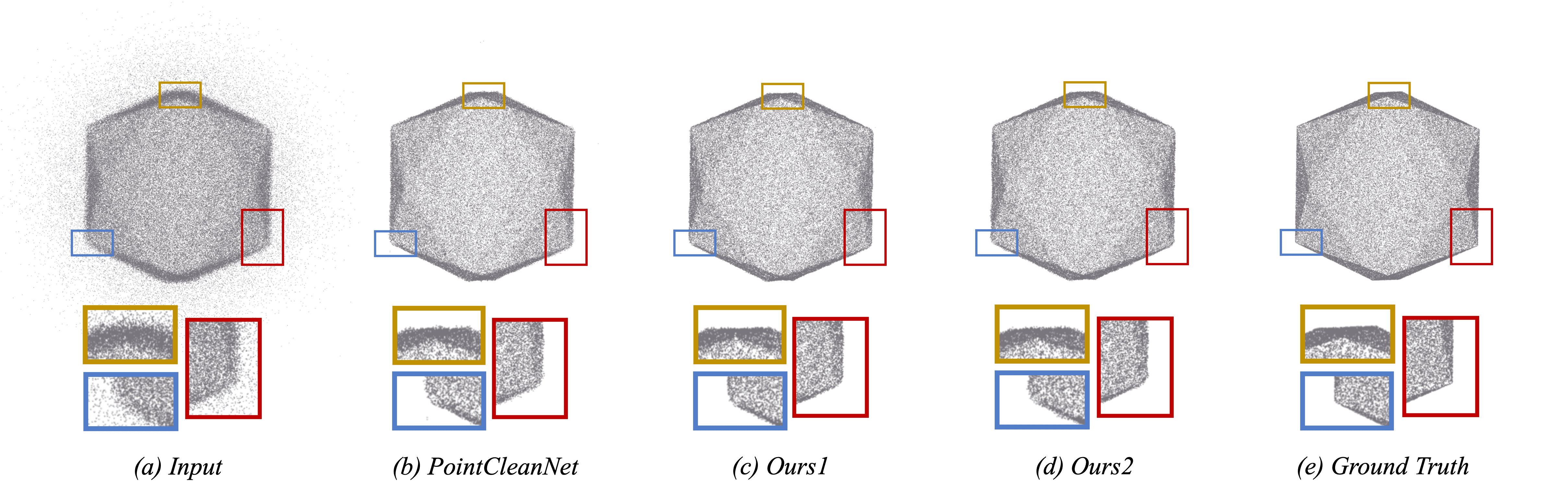}}
\caption{Experiments of qualitative results of the overall model for the icosahedron data. (a) Point clouds contaminated by outliers and noise. (b) Processing results of conventional method. (c) Processing results of proposed method1 (Ours1). (d) Processing results of proposed method2 (Ours2). (e) Ground truth.}
\label{fig}
\end{figure*}

\subsection{Overall Model Evaluation}
To evaluate the performance of the overall model, which consists of \textit{Outlier Detector} and \textit{Denoiser}. The data is applied to a contaminated icosahedron. It is subjected to gaussian noise with a standard deviation of $20$\% on the diagonal of the shape bounding box. In addition, $30$\% of the points farther from the surface than $1.5$\% standard deviation are converted as outliers.

Quantitative results are shown in TABLE I\hspace{-1.2pt}V. It shows that the "Ours$1$" is the best. In particular, it can be seen that for \textit{PointCleanNet}, the chamfer distance was reduced by $0.53\times10^{-4}$. It can also be seen that the "Ours$2$" is slightly inferior to "Ours$1$", but performs equally well. In addition, it can reduce the computation time by $13.3$\%, showing that two proposed methods are efficient.

Qualitative results are also shown in Fig. 4. It can be seen that the two proposed methods detect outliers at a higher gaussian level than the conventional methods. It shows that they can reduce the unnatural distortion of the plane and smoothing of the edge areas.
These results confirm the effectiveness of the two proposed methods. From the above, "Ours${1}$ outperforms the conventional method in terms of performance. In addition, "Ours${2}$ shows that it is more efficient than the method that computes the graph dynamically, in addition to having the same performance as the method that computes the graph dynamically.
\vspace{1.5mm}
\section{Conclusion}
In this paper, we proposed new method, point cloud denoising and outlier detection. Specifically, we introduced two types of graph convolutional layer based on \textit{Dynamic Graph CNN} to \textit{PointCleanNet}. Experiments show that the two proposed models are superior in AUPR and chamfer distance. Thus, the effectiveness of the two models was confirmed. In particular, we found that the "Ours$1$" is accuracy specific, while the proposed "Ours$2$" is balance between accuracy and computation speed. Depending on the issue to be applied, it is necessary to be flexible in pursuit of accuracy or in consideration of calculation speed.


\vspace{0.5mm}
\begin{thebibliography}{00}
\bibitem{b1} J. Digne and C. Franchis, "The Bilateral Filter for Point Clouds," Image Processing On Line (IPOL), pp.278-287, Jul. 2017.
\vspace{1.05mm}
\bibitem{b2} M. Alexa, J Behr, D. Cohen-Or S. Fleishman, D. Levin, C.T. Silva, “Point Set Sur faces,” VIS '01: Proceedings of the conference on Visualization ‘01, pp.21-28, Oct. 2001.
\vspace{1.05mm}
\bibitem{b3} M.J. Rakotosaona, V .L. Barbera, P . Guerrero, N.J. Mitra, and M. Ovsjanikov, “PointCleanNet: Learning to denoise and remove outliers from dense point clouds,” In Computer Graphics Forum, vol.39, pp.185-203, Wiley Online Library, 2020.
\vspace{1.05mm}
\bibitem{b4} Y. Wang, Y. Sun, Z. Liu, S.E. Sarma, M.M. Bronstein, and J.M. Solomon, “Dynamic Graph CNN for Learning on Point Clouds,” ACM Transactions on Graphics (TOG), vol.38, Jun. 2019.
\vspace{1.05mm}
\bibitem{b5} Y. Li, H. Chen, Z. Cui, R. Timofte, M. Pollefeys, G. Chirikjian, and L.V. Gool, “Towards Efficient Graph Convolutional Networks for Point Cloud Handling,” 2021 IEEE/CVF International Conference on Computer Vision (ICCV), pp.3732-3742, Apr. 2021.
\vspace{1.05mm}
\bibitem{b6} K. He, X. Zhang, S. Ren, J. Sun, “Delving Deep into Rectifiers: Surpassing Human-Level Performance on ImageNet Classification,” 2015 IEEE International Conference on Computer Vision (ICCV), pp.1026-1034, Dec. 2015.
\end{thebibliography}
\end{document}